# An Anchor-Free Region Proposal Network for Faster R-CNN based Text Detection Approaches

Zhuoyao Zhong[1,2,*], Lei Sun[2], Qiang Huo[2]
[1]School of EIE., South China University of Technology, Guangzhou, China
[2]Microsoft Research Asia, Beijing, China
zhuoyao.zhong@gmail.com, lsun@microsoft.com, qianghuo@microsoft.com

## Abstract

The anchor mechanism of Faster R-CNN and SSD framework is considered not effective enough to scene text detection, which can be attributed to its IoU based matching criterion between anchors and ground-truth boxes. In order to better enclose scene text instances of various shapes, it requires to design anchors of various scales, aspect ratios and even orientations manually, which makes anchor-based methods sophisticated and inefficient. In this paper, we propose a novel anchor-free region proposal network (AF-RPN) to replace the original anchor-based RPN in the Faster R-CNN framework to address the above problem. Compared with a vanilla RPN and FPN-RPN, AF-RPN can get rid of complicated anchor design and achieve higher recall rate on large-scale COCO-Text dataset. Owing to the high-quality text proposals, our Faster R-CNN based two-stage text detection approach achieves state-of-the-art results on ICDAR-2017 MLT, ICDAR-2015 and ICDAR-2013 text detection benchmarks when using single-scale and single-model (ResNet50) testing only.

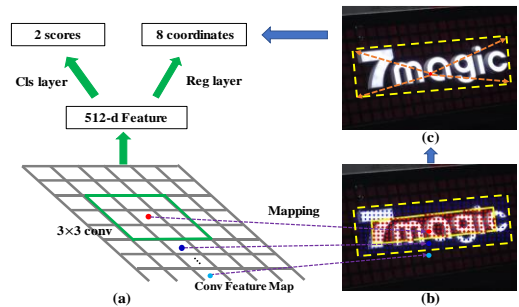

Figure 1. (a) A detection module of AF-RPN; (b) Mapping each pixel in the conv feature map to the corresponding sliding point in the raw image; Examples of text (red), ignored (blue) and non-text (outside the text region) sliding points; (c) Direct regression from a text sliding point to the four vertices of a ground truth box.

## 1 Introduction

Scene text detection has drawn considerable attentions of computer vision and document analysis communities recently [Shahab et al., 2011; Karatzas et al., 2013; Karatzas et al., 2015] owing to the increasing demands for many content-based visual intelligent applications, e.g., image and video retrieval, scene understanding and target geolocation. However, because of diverse text variabilities in colors, fonts, orientations, languages and scales, extremely complex and text-like backgrounds, as well as some distortions and artifacts caused by image capturing like non-uniform illumination, low contrast, low resolution and occlusion, text detection in natural scene images is still an unsolved problem.

Nowadays, with the astonishing development of deep learning, state-of-the-art convolutional neural network (CNN) based object detection frameworks such as Faster R-CNN [Ren et al., 2015] and SSD [Liu et al., 2016] have been widely used to address the text detection problem and outperformed substantially traditional MSER [Matas et al., 2002] or SWT [Epshtein et al., 2010] based bottom-up text detection approaches. However, Faster R-CNN and SSD are found to be not flexible enough for text detection because of their anchor (called default box in SSD) mechanism [He et al., 2017]. Anchors are used as reference boxes in both Faster R-CNN and SSD to predict corresponding region proposals or objects, and the label of each anchor is determined by its Intersection-over-Union (IoU) overlap with ground-truth bounding boxes [Ren et al., 2015]. If we want an object to be detected, there should be at least one anchor which has high enough IoU overlap with this object. So, to achieve high recall, anchors with various scales and shapes should be designed to cover the scale and shape variabilities of objects in images. As scene text instances have wider variability in scales, aspect ratios and especially orientations than general objects, it requires much more complicated anchor design, i.e., more scales, aspect ratios and orientations [Zhong et al., 2017; Liao et al., 2016; Ma et al., 2017; Liu and Jin, 2017], which makes anchor-based methods sophisticated and inefficient. Recently, the idea of DenseBox [Huang et al., 2015] is borrowed to overcome this problem in some text detection methods [He et al., 2017; Zhou et al., 2017], which use a fully convolutional neural network (FCN) [Long et al., 2015] to directly output the pixel-wise textness scores and bounding boxes of the concerned text instances through all locations and scales of an

---

image. Although more flexible, the capabilities of these approaches are affected. For example, they cannot detect long or large text instances robustly, which occur very often in "Multilingual scene text detection" scenarios [Nayef et al., 2017], as the maximal size of text instances that can be handled by the detector is limited by the receptive field (*RF*) sizes of the used convolutional feature maps.

To overcome the above problems, we propose to incorporate the smart "anchor-free" idea of DenseBox into the Faster R-CNN framework. Specifically, we propose a novel anchor-free region proposal network (AF-RPN) to replace the original anchor-based RPN so that our text detector can possess high flexibility and high capability at the same time. As illustrated in Fig. 1, each pixel in a specific convolutional feature map can be mapped to a point (called a sliding point hereinafter) in the raw image. For each sliding point that locates in a text core region (red points in Fig. 1 (b)), AF-RPN directly predicts the offsets from it to the bounding box vertices of the concerned text instance (Fig. 1 (c)). In this way, AF-RPN can generate high-quality inclined text proposals directly in an anchor-free manner, which can get rid of complicated hand-crafted anchor design. Moreover, the label definitions for sliding points in AF-RPN are much easier than IoU based label definitions for anchors in the original RPN, where we only need to determine whether a sliding point is inside any ground-truth bounding box's core region. Compared with DenseBox based text detectors, Faster R-CNN based text detectors can deal with long or large text instances much more effectively. This is because the ROI pooling algorithm in the second-stage Fast R-CNN can enlarge the *RF* size of pooled features for each proposal significantly, which can improve not only the bounding box regression precision of long text instances, but also the text/non-text classification accuracy. Furthermore, unlike DenseBox, we let AF-RPN extract text proposals from multi-scale feature maps of feature pyramid network (FPN) [Lin et al., 2017] in a scale-friendly manner so that AF-RPN can be more robust to text scales. Thanks to this, our text detector can achieve superior text detection performance with only single-scale testing.

Extensive experiments demonstrate that, as a new region proposal generation approach, AF-RPN can achieve higher recall rate than the vanilla RPN [Ren et al., 2015] and FPN-RPN [Lin et al., 2017] on the large-scale COCO-Text [Veit et al., 2016] dataset. Owing to the high-quality text proposals, our Faster R-CNN based two-stage text detection approach, i.e., AF-RPN + Fast R-CNN, achieves state-of-the-art results on ICDAR-2017 MLT [Nayef et al., 2017], ICDAR-2015 [Karatzas et al., 2015] and ICDAR-2013 [Karatzas et al., 2013] text detection benchmark tasks.

## 2 Related Work

Existing text detection methods can be roughly divided into two categories: bottom-up [Neumann and Matas, 2010; Epshtein et al., 2010; Yao et al., 2012; Yin et al., 2014; Yin et al., 2015; Sun et al., 2015] and top-down methods [Jaderberg et al., 2014; Zhang et al., 2016; Yao et al., 2016; Jaderberg et al., 2016; Gupta et al., 2016; Ma et al., 2017; Liu and Jin, 2017; Zhou et al., 2017; He et al., 2016]. Bottom-up methods are generally composed of three major steps, i.e., candidate text connected components (CCs) extraction (e.g., based on MSER [Matas et al., 2002] or SWT [Epshtein et al., 2010]), text/non-text classification and text-line grouping. These methods, especially MSER based ones, were once the mainstream methods before the deep learning era. However, recently these methods fall behind of CNN based top-down approaches in terms of both accuracy and adaptability, especially when dealing with more challenging "Incidental Scene Text" [Karatzas et al., 2015] and "Multi-lingual Scene Text Detection" [Nayef et al., 2017] scenarios.

So, CNN based top-down approaches have become the mainstream recently. Some works [Zhang et al., 2016; Yao et al., 2016] borrowed the idea of semantic segmentation and employed FCN [Long et al., 2015] to solve text detection problem. As only coarse text-blocks can be extracted from the text saliency map predicted by FCN, post-processing steps are needed to extract accurate bounding boxes of text-lines. Instead of merely predicting text saliency map with FCN, some works [Zhou et al., 2017; He et al., 2017] borrowed the idea of DenseBox [Huang et al., 2015] and used a one-stage FCN to output pixel-wise textness scores as well as the quadrilateral bounding boxes through all locations and scales of an image. Although these approaches are more flexible, they cannot handle long or large text instances effectively and efficiently. Object detection frameworks are also widely used to address the text detection problem. The work by [Jaderberg et al., 2016] adapted R-CNN [Girshick et al., 2014] for text detection. The performance of this approach is limited by the traditional region proposal generation methods. The work by [Gupta et al., 2016] resembles YOLO framework [Redmon, 2016] and employed a fully-convolutional regression network to perform text detection. The works by [Zhong et al., 2017] and [Liao et al., 2016] employed anchor-based Faster R-CNN [Ren et al., 2015] and SSD [Liu et al., 2016] frameworks to solve horizontal text detection problem respectively. In order to extend Faster R-CNN and SSD to multi-oriented text detection, the works by [Ma et al., 2017; Liu and Jin, 2017] proposed quadrilateral anchors to hunt for inclined text proposals which can better fit the multi-oriented text instances. However, as mentioned above, these anchor-based methods are not effective and flexible enough for text detection, which lead to inferior performance. In this paper, we propose a novel AF-RPN to improve the flexibility and capability of Faster R-CNN, which makes our approach achieve superior performance on various text detection benchmarks.

## 3 Anchor-free Region Proposal Network

Our proposed AF-RPN is composed of a backbone network and three scale-specific detection modules. The backbone network is responsible for computing a multi-scale convolutional feature pyramid over the full input image. Three detection modules are attached on different pyramid levels, and designed to detect small, medium and large text instances, respectively. Each detection module contains a small network with two sibling output layers for text/non-text classification and quadrilateral bounding box regression respectively. A schematic view of our AF-RPN model is depicted in Fig. 2 and details are described in the following subsections.

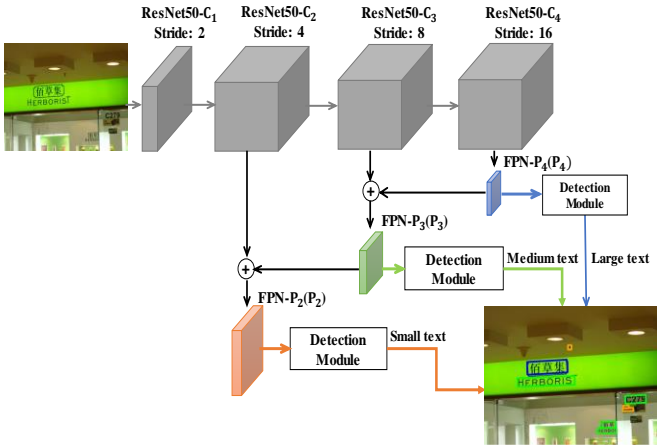

Figure 2. Architecture of the proposed AF-RPN, which consists of an FPN backbone network and three scale-specific detection modules (Fig.1 (a)) for small, medium and large text detection, respectively. Visualization of text proposals after score thresholding and NMS.

### 3.1 Network Architecture of AF-RPN

We adopt FPN [Lin et al., 2017] as the backbone network for AF-RPN. In brief, FPN enhances a standard convolutional network with a top-down pathway and lateral connections to construct a rich and multi-scale feature pyramid from a single resolution input image. Each level of the pyramid can be effectively used for detecting objects of scales within a specific range. Following [Lin et al., 2017], we build FPN on top of the ResNet50 architecture [He et al., 2016], and construct a feature pyramid with three levels, i.e., $P_2$, $P_3$ and $P_4$, whose strides are 4, 8, 16, respectively. Implementations of FPN generally follow [Lin et al., 2017] with just two modest differences. First, we do not include the pyramid level $P_5$ as its resolution is too coarse for text detection task. Second, all feature pyramid levels have $C = 512$ channels instead of 256. We refer readers to [Lin et al., 2017] for further details.

Three scale-specific detection modules are attached to $P_2$, $P_3$ and $P_4$, respectively. Similar to RPN [Ren et al., 2015], each detection module can be considered as a sliding-window detector, which uses a small network to perform text/non-text classification and bounding box regression in each $3 \times 3$ sliding window on a single-scale pyramid level (Fig. 1(a)). The small network is implemented as a $3 \times 3$ convolutional layer followed by two sibling $1 \times 1$ convolutional layers, which are used for predicting textness score and bounding box coordinates respectively. We propose a scale-friendly learning method to learn three detection modules that are designed to detect small, medium, and large text instances respectively. This can effectively relieve the learning difficulties of text/non-text classification and bounding box regression of each detection module, thus make AF-RPN be able to deal with large text-scale variance more robustly. The details of scale division are described in Section 3.2, and the ground-truth definition of AF-RPN is elaborated in Section 3.3.

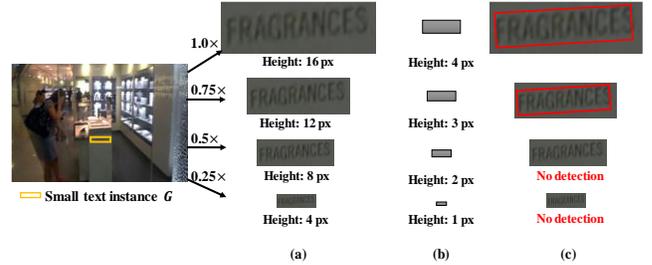

Figure 3. (a) Down-sampling the input image by a factor of 1.0, 0.75, 0.50 and 0.25 such that the shorter side of the small text instance $G$ has 16, 12, 8 and 4 px, respectively. (b) The corresponding features of $G$ with different sizes on the pyramid level $P_2$. (c) The corresponding detection results of $G$ with different sizes by using a detection module of AF-RPN attached on $P_2$.

### 3.2 Scale-friendly Learning

In the training stage, we assign text instances to the three detection modules based on the spatial sizes of their features on the corresponding pyramid levels. An interesting observation is that when the shorter side of the features for a text instance on a pyramid level is less than 3 pixels (px), the performance of the corresponding detection module degrades dramatically. Examples can be seen in Fig. 3. The height (shorter side) of the text instance $G$ enclosed by a yellow bounding box in the raw image is 16 px. When the raw image is down-sampled by a scale of 0.75, 0.50 and 0.25, the height of $G$ is decreased to 12, 8 and 4 px (Fig. 3 (a)), and the height of $G$' features on $P_2$ is reduced to 3, 2, 1 px (Fig. 3(b)), respectively. Then we use a detection module of AF-RPN attached on $P_2$, which is trained to detect text instances of scales (shorter sides) in the range of [4px, 24px], to detect this text instance on these 4 images. From Fig. 3(c), we find that no effective proposals can be detected when its height is less than 12 px in the input image even if our training set contains lots of these small text instances. This must be caused by the insufficient features of $G$ whose shorter sides have less than 3 px. Based on this observation, when assigning a text instance to a pyramid level, we ensure that the shorter side of its features on this pyramid level has no less than 3 px. As the strides of $P_2$, $P_3$ and $P_4$ are 4, 8, 16 px, the shorter side of text instances assigned to them should have no less than 12, 24 and 48 px respectively. Consequently, we empirically classify text instances into three groups according to their shorter side lengths, i.e., small text (4px-24px), medium text (24px – 48px) and large text (> 48px).

### 3.3 Label Generation

Text instances in text detection tasks are usually labeled in word-level with quadrilateral or axis-aligned bounding boxes. To ease implementation, for quadrilateral bounding boxes, we use their minimum enclosing boxes (oriented rectangles) as new ground-truth bounding boxes (yellow dashed lines in Fig. 1 (b)). It is inevitable that some surrounding backgrounds can be included in the ground-truth bounding boxes when they are not tight enough. To reduce the influence of background noise on text/non-text classification, following [Zhou et al., 2017; He et al., 2017], we shrink the short and long sides of each ground-truth rectangle by scaling factors of 0.5 and 0.8 respectively to create the corresponding core text region (yellow solid lines in Fig. 1 (b)), and only sliding points within core regions are taken as positive. Sliding points outside core regions but inside ground-truth rectangles are assigned a "DON'T CARE" label, and are ignored during training (Fig. 1(b)). Sliding points outside all ground-truth rectangles are taken as negative. For each positive sliding point, we predict the coordinates of its bounding box directly. Let $p_t = (x_t, y_t)$ denote a positive sliding point, which is located in a ground-truth rectangle $G$. Let $\{p_i = (x_i, y_i) | i \in \{1,2,3,4\}\}$ denote the vertices of $G$. Then the coordinate offsets from $p_t$ to $G$'s vertices can be denoted as $\{\Delta_i = (\Delta_{x_i}, \Delta_{y_i}) | i \in \{1,2,3,4\}\}$, where $\Delta_{x_i} = (x_i - x_t)$ and $\Delta_{y_i} = (y_i - y_t)$ (Fig. 1 (c)). Considering the fact that the numerical ranges of $\Delta_{x_i}$ and $\Delta_{y_i}$ are very large, we normalize them as follows: $\Delta_{x_i} = (x_i - x_t)/norm$, $\Delta_{y_i} = (y_i - y_t)/norm$, where $norm$ represents the normalization term. If $p_t$ is on $P_2$ or $P_3$, $norm$ is set as the upper bound of the corresponding scale range, i.e., $norm = 24$ or $norm = 48$. If $p_t$ is on $P_4$, $norm$ is set as a proportion of the $RF$ of $P_4$ (related to $3 \times 3$ units), i.e., $norm = \alpha RF_{P_4}$, where $\alpha = 0.5$.

# 4 Light Head Faster R-CNN with AF-RPN

After region proposal generation step, the top-$N_1$ scoring detection results of each detection module in AF-RPN are selected respectively to construct a proposal set $\{P\}$. Then, we use standard NMS algorithm (IoU = 0.7) to remove redundant proposals in $\{P\}$, and select the top-$N_2$ scoring proposals for Fast R-CNN. $N_1$, $N_2$ are set to 2000 and 300 in the training and testing stage, respectively. Next, we adopt the same scale division criterion in Section 3.2 to classify text proposals into small, medium and large text proposal groups, which are assigned to pyramid level $P_2$, $P_3$ and $P_4$, respectively. Similar to AF-RPN, three Fast R-CNN detectors, which do not share parameters, are assigned to pyramid level $P_2$, $P_3$ and $P_4$, respectively. The head design of each Fast R-CNN detector follows the idea of Light Head R-CNN [Li et al., 2017]. Concretely, we apply two large separable convolution layers (i.e., $15 \times 1$ and $1 \times 15$ convolutions) on specific pyramid level to create a large $RF$ size and thin (490 channels) feature maps, from which we adopt PS ROI pooling [Dai et al., 2015] to extract $7 \times 7$ features for each proposal, and then feed the pooled features into a single fully-connected layer with 2048 units before the final text/non-text classification and bounding box regression layers.

# 5 Training

## 5.1 Loss Functions

**Multi-task loss for AF-RPN.** There are two sibling output layers for each scale-specific detection module, i.e., text/non-text classification layer and a quadrilateral bounding box regression layer. The multi-task loss function for each detection module is denoted as follows:
$$L(c, c^*, t, t^*) = \lambda_{cls} L_{cls}(c, c^*) + \lambda_{loc} L_{loc}(t, t^*), \quad (1)$$

where $c$ and $c^*$ are predicted and ground-truth labels for each sliding point respectively, $L_{cls}(c, c^*)$ is a softmax loss for classification task; $t$ and $t^*$ represent the predicted and ground-truth 8-dimensional normalized coordinate offsets from $p_t$ to $G$, $L_{loc}(t, t^*)$ is a smooth-L$_1$ loss [Girshick, 2015] for regression task. $\lambda_{cls}$ and $\lambda_{loc}$ are two loss-balancing parameters, and we set $\lambda_{cls} = 1$, $\lambda_{loc} = 3$.

The total loss of AF-RPN is the sum of the losses of three scale-specific detection module.

**Multi-task loss for Fast R-CNN.** The loss function for Fast R-CNN is the same as Equation (1). Only the parameters $\lambda_{cls}$ and $\lambda_{loc}$ are set differently. Here, we set $\lambda_{cls} = 1$, $\lambda_{loc} = 1$. Compared to AF-RPN, there are some differences in coordinate offsets normalization for regression task. Let $P$ be the input proposal and $(P_x, P_y, P_w, P_h)$ be the center coordinates, height and width of its axis-aligned bounding box respectively. We directly use $P_w$ and $P_h$ to normalize the coordinate offsets from $(P_x, P_y)$ to $G$'s vertices as follows: $\Delta_{x_i} = (x_i - P_x)/P_w$, $\Delta_{y_i} = (y_i - P_y)/P_h$, where $i \in \{1,2,3,4\}$. The total loss of Fast R-CNN is the sum of losses of three individual Fast R-CNN detectors.

## 5.2 Training Details

In each training iteration of AF-RPN, we sample a mini-batch of 128 positive and 128 negative sliding points for each detection module. Similarly, for Fast R-CNN, we sample a mini-batch of 64 positive and 64 negative text proposals for each Fast R-CNN detector. A proposal is assigned a positive label if it has an IoU overlap higher than 0.5 with any ground-truth bounding box, and a negative label if its IoU overlap is less than 0.3 for all ground-truth bounding boxes. For efficiency, the IoU overlaps between proposals and ground-truth boxes are calculated using their axis-aligned rectangular bounding boxes. Note that each ground-truth bounding box is assigned to only one detection module of AF-RPN or one Fast R-CNN detector, and ignored by other two. Moreover, we apply online hard example mining (OHEM) as [Dai et al., 2015] to effectively suppress hard text-like background examples during the training phase of Fast R-CNN.

# 6 Experiments

## 6.1 Datasets and Evaluation Protocols

We evaluate our approach on several standard benchmark tasks, including ICDAR-2017 MLT [Nayef et al., 2017] and ICDAR-2015 [Karatzas et al., 2015] for multi-oriented text detection, ICDAR-2013 [Karatzas et al., 2013] for horizontal text detection, and COCO-Text [Veit et al., 2016] for proposal quality evaluation.

To make our results comparable to others, we use the online official evaluation tools to evaluate the performance of our approach on ICDAR-2017 MLT, ICDAR-2015 and ICDAR-2013 testing sets, which contains 9,000, 500 and 233 images, respectively. We use recall rate as an evaluation metric to compare the performance of different region proposal approaches on COCO-Text validation set (10,000 images).

Table 1. Region proposal quality evaluation on COCO-Text validation set (%). The column "feature" denotes the feature maps on which the prediction layers are attached. The column "#anchor/#sp" represents the number of anchors or sliding points used during inference for anchor-based RPN/FPN-RPN and AF-RPN respectively.

| Method | feature | #anchor/#sp | $R_{50}^{.5}$ | $R_{50}^{.75}$ | $AR_{50}^{.5:.05:.95}$ | $R_{100}^{.5}$ | $R_{100}^{.75}$ | $AR_{100}^{.5:.05:.95}$ | $R_{300}^{.5}$ | $R_{300}^{.75}$ | $AR_{300}^{.5:.05:.95}$ |
|---|---|---|---|---|---|---|---|---|---|---|---|
| (a) RPN | $C_4$ | 45k | 67.2 | 22.8 | 30.6 | 76.9 | 27.9 | 35.9 | 86.6 | 33.8 | 41.7 |
| (b) FPN-RPN | $\{P_k\}$ | 315k | 67.5 | 28.8 | 33.5 | 77.2 | 36.0 | 39.8 | 87.4 | 47.2 | 48.0 |
| (c) FPN-RPN | $P_2$ | 720k | 67.7 | 30.1 | 34.0 | 76.4 | 36.7 | 39.6 | 83.6 | 43.8 | 44.9 |
| (d) AF-RPN | $\{P_k\}$ | 52.5k | **73.3** | **35.0** | **38.2** | **81.8** | **41.3** | **43.6** | **89.3** | **48.2** | **49.2** |

### 6.2 Implementation Details

The weights of Resnet50 related layers in the FPN backbone network are initialized by using a pre-trained Resnet50 model for ImageNet classification [He et al., 2016]. The weights of other new layers for FPN, AF-RPN and Fast R-CNN are initialized by using random weights with Gaussian distribution of 0 mean and 0.01 standard deviation. During training stage, we train the AF-RPN model until convergence firstly. Then we use this well-trained AF-RPN model to initialize the Faster R-CNN model, which is then fine-tuned with the approximate end-to-end training algorithm. All the models are optimized by the standard SGD algorithm with a momentum of 0.9 and weight decay of 0.0005.

The number of training iterations and adjustment strategy of learning rate depend on the size of different datasets. Specifically, for ICDAR-2017 MLT, we use the training and validation data, i.e., a total of 9,000 images for training. Both AF-RPN and Faster R-CNN models are trained for 200K iterations with the initial learning rate of 0.001, which is then divided by 10 at 90K and 180K iterations. As the training sets of ICDAR-2015 and ICDAR-2013 are too small, we directly use the model trained on ICDAR-2017 MLT as pre-trained model, which is then fine-tuned on the training set of ICDAR-2015 and ICDAR-2013 respectively. All models for these two datasets are trained for 40K iterations with the initial learning rate of 0.0005, which is divided by 5 at 20K iterations. For COCO-Text, we train the models on its training set and report experimental results on validation set. All the models are trained for 250K iterations with the initial learning rate of 0.001, which is then divided by 10 at 100K and 200K iterations. Our experiments are conducted on Caffe framework [Jia et al., 2014].

### 6.3 Region Proposal Quality Evaluation

A series of ablation experiments are conducted to compare AF-RPN to RPN and FPN-RPN in region proposal generation task on the large-scale COCO-Text dataset. For fair comparison, we evaluate the axis-aligned rectangular proposal quality because the original RPN and FPN-RPN cannot output quadrilateral proposal without oriented anchors design. We compute the recall rates $R_\#^{.5}$ and $R_\#^{.75}$ at a single IoU threshold of 0.5 and 0.75, respectively, while also compute the average recall rate $AR_\#^{.5:.05:.95}$ at multiple IoU thresholds between 0.50 and 0.95 with an interval of 0.05, when using a given fixed number (#) of text proposals. We report results for 50, 100 and 300 proposals per image (300 proposals are used for Fast R-CNN in testing stage). The scale $S$ (the length of the shorter side of an image) for testing image is set as 800 for all experiments in this section.

Following [Zhong et al., 2017], we design a complicated set of anchors for RPN and FPN-RPN. Specifically, for RPN, we use 3 scales {32, 64, 128} and 6 aspect ratios {0.2, 0.5, 0.8, 1.0, 1.2, 1.5}, i.e., 18 anchors, at each sliding position on $C_4$. We train two models for FPN-RPN. For the first one, we use 6 aspect ratios and a single scale in {32, 64, 128}, i.e., 6 anchors, at each position on each pyramid level in $\{P_2, P_3, P_4\}$. The second one is similar to RPN and we assign all anchors on $P_2$. In the training stage of RPN and FPN-RPN, an anchor is assigned a positive label if it has an IoU overlap higher than 0.5 with any ground-truth boxes or the highest IoU for a given ground-truth box and a negative label if it has an IoU less than 0.1 for all ground-truth boxes as [Zhong et al., 2017]. The training strategies are kept the same as AF-RPN. The results are listed in Table 1. It can be seen that our proposed AF-RPN outperforms RPN and FPN-RPN substantially in all evaluation metrics. When the evaluated number of proposals drops from 300 to 50, the improvements are much more significant, which demonstrates the effectiveness of our proposed AF-RPN.

Moreover, AF-RPN can directly output inclined text proposals for oriented text instances. For quadrilateral proposal quality evaluation, AF-RPN achieves 88.9%, 92.2% and 95.4% in $R_{50}^{.5}$, $R_{100}^{.5}$ and $R_{300}^{.5}$ on ICDAR-2015, respectively.

### 6.4 Text Detection Performance Evaluation

In this section, we evaluate the performance of our proposed Faster R-CNN based text detection approach on various benchmarks. We use the top-300 scoring text proposals generated by AF-RPN for the succeeding Fast R-CNN. Detection results from different Fast R-CNN detectors are aggregated with Skewed NMS [Ma et al., 2017]. All the experiments are based on single-model and single-scale testing. The scale $S$ of testing image is set as 512, 800 and 1280 for ICDAR-2013, ICDAR-2015 and ICDAR-2017 MLT respectively.

**ICDAR-2017 MLT.** As this dataset is built for ICDAR-2017 Robust Reading Competition and has been released recently, there is no result reported in published literature. Thus, we collect competition results [Nayef et al., 2017] for comprehensive comparisons. As shown in Table 2, our approach outperforms the top-1 competition result remarkably by improving F-measure from 0.65 to 0.70. Considering that ICDAR-

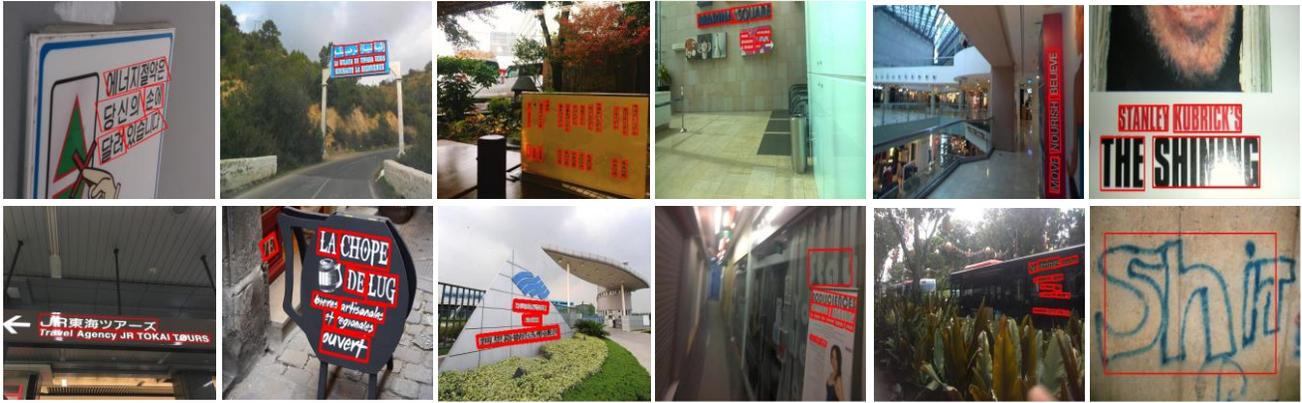

Figure 4. Detection results of our approach: 1st-3rd col: ICDAR-2017 MLT; 4th-5th col: ICDAR-2015; 6th col: ICDAR-2013.

Table 2. Comparison with prior arts on ICDAR-2017 MLT.

| Method | Recall | Precision | F-measure |
|---|---|---|---|
| linkage-ER-Flow | 0.26 | 0.44 | 0.32 |
| TH-DL | 0.35 | 0.68 | 0.46 |
| SARI_FDU_RRPN_v2 | 0.55 | 0.67 | 0.61 |
| SARI_FDU_RRPN_v1 | 0.56 | 0.71 | 0.62 |
| Sensetime OCR | **0.69** | 0.57 | 0.63 |
| SCUT_DLVClab1 | 0.55 | **0.80** | 0.65 |
| Proposed | 0.66 | 0.75 | **0.70** |

Table 3. Comparison with prior arts on ICDAR-2015. MS means multi-scale testing.

| Method | Recall | Precision | F-measure |
|---|---|---|---|
| [Liu and Jin, 2017] | 0.68 | 0.73 | 0.71 |
| [Shi et al., 2017] | 0.77 | 0.73 | 0.75 |
| [Ma et al., 2017] | 0.73 | 0.82 | 0.77 |
| [Han et al., 2017]+MS | 0.77 | 0.79 | 0.78 |
| [Zhou et al., 2017] (PVANET2x+MS) | 0.78 | 0.83 | 0.81 |
| [Zhou et al.,2017] (ResNet50) | 0.77 | 0.85 | 0.81 |
| [He et al., 2017]+MS | 0.80 | 0.82 | 0.81 |
| Proposed | **0.83** | **0.89** | **0.86** |

2017 MLT is an extremely challenging and the first large-scale multi-lingual text detection dataset, the superior performance achieved by our proposed approach can demonstrate apparently its advantage.

**ICDAR-2015.** On the challenging ICDAR-2015 task, as shown in Table 3, our approach achieves the best results of 0.83, 0.89 and 0.86 in recall, precision and F-measure respectively, outperforming other recently published CNN-based approaches substantially.

**ICDAR-2013.** We also evaluate our approach on ICDAR-2013, which is a popular dataset for horizontal text detection. As shown in Table 4, although ICDAR-2013 is well-tuned by many previous methods, our approach still achieves the best result of 0.92 F-measure. In terms of run-time, our approach takes about 0.50s on average for each 512×1024 image when using a single M40 GPU.

Table 4. Comparison with prior arts on ICDAR-2013. MS means multi-scale testing.

| Method | Recall | Precision | F-measure |
|---|---|---|---|
| [Gupta et al., 2016]+MS | 0.76 | 0.92 | 0.83 |
| [Shi et al., 2017] | 0.83 | 0.88 | 0.85 |
| [Liao et al., 2016]+MS | 0.83 | 0.89 | 0.86 |
| [He et al., 2017]+MS | 0.81 | 0.92 | 0.86 |
| [Tian et al., 2016] | 0.83 | 0.93 | 0.88 |
| [Zhou et al., 2017] (PVANET2x) | 0.83 | 0.93 | 0.88 |
| [Han et al., 2017]+MS | 0.88 | 0.93 | 0.90 |
| Proposed | **0.90** | **0.94** | **0.92** |

**Qualitative Results.** The superior performance achieved on the above three datasets shows the effectiveness and robustness of our approach. As shown in Fig. 4, our text detector can detect scene text regions under various challenging conditions, such as low-resolution, non-uniform illumination, large aspect ratios as well as varying orientation. However, our approach still has some limitations. First, our approach cannot deal with curved text instances. Second, our approach still struggles with extremely small text instances whose shorter sides are less than 8 pixels in resized images. More researches are needed to address these challenging problems.

### 6.5 Comparisons with Prior Arts

Here we compare with other relevant scene text detection methods for better understanding the superiority of our approach. Works in [Liu and Jin, 2017] and [Ma et al., 2017] introduce oriented anchor strategy for one-stage SSD and two-stage Faster R-CNN framework. The performances of these anchor-based methods are obviously inferior to our anchor-free approach, i.e., 0.71 and 0.78 vs. 0.86 in F-measure on ICDAR-2015. In addition, although DenseBox based text detection methods [Zhou et al., 2017; He et al., 2017] also make use of the "anchor-free" concept, their capability is limited for large or long text as shown in the first row of Fig. 5. Our Faster R-CNN based two-stage approach, i.e., AF-RPN + Fast R-CNN, can overcome this limitation effectively, thus improve the F-measure from 0.81 to 0.86 on ICDAR-2015.

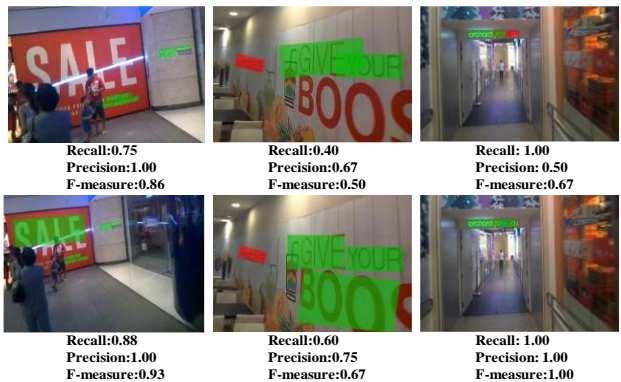

Figure 5. Qualitative detection results of (Zhou et al., 2017)'s DenseBox based text detector with ResNet50 backbone network obtained from http://east.zxytim.com/ (1st row) and our approach (2nd row) on ICDAR-2015 dataset. Green, red, grey regions represent correctly detected, wrongly detected, ignored text regions, respectively. Visualizations are captured from the online evaluation system (http://rrc.cvc.uab.es/?ch=4).

Some qualitative comparison results are shown in the second row of Fig. 5.

## 7 Conclusion and Discussion

In this paper, we present AF-RPN as an anchor-free and scale-friendly region proposal network for the Faster R-CNN framework. Comprehensive comparisons with RPN, FPN-RPN on the large-scale COCO-Text datasets demonstrate the superior performance of our proposed AF-RPN used as a new region proposal network. Owing to the high-quality text proposals, our Faster R-CNN based text detector, i.e., AF-RPN + Fast R-CNN, achieves state-of-the-art results on ICDAR-2017 MLT, ICDAR-2015 and ICDAR-2013 text detection benchmarks based on single-scale and single-model testing. Future direction is to explore the effectiveness of the proposed AF-RPN in other detection tasks, such as generic object detection, face detection, and pedestrian detection, etc.